\documentclass[runningheads]{llncs}
\usepackage{graphicx}
\newcommand{\orcid}[1]{
	\href{https://orcid.org/#1}{\includegraphics[scale=0.4]{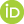}}
}
\usepackage{fontawesome}  
\usepackage{cite}
\usepackage{amsmath,amssymb,amsfonts}

\usepackage[ruled, vlined, linesnumbered]{algorithm2e}

\SetCommentSty{mycommfont}
\usepackage{algpseudocode}
\usepackage{graphicx}
\usepackage{textcomp}
\usepackage{xcolor}
\usepackage{color,soul}
\usepackage{float}
\usepackage{nicefrac}
\usepackage{hyperref}

\newcommand{\mechanism}{\mathcal{M}}
\newcommand{\universeLog}{\mathcal{L}}
\newcommand{\universeActivity}{\mathcal{A}}

\newcommand{\multiset}{\mathcal{B}}

\begin{document}
\title{Releasing Differentially Private Event Logs Using Generative Models\thanks{\scriptsize Funded under the Excellence Strategy of the Federal Government and the L{\"a}nder. We also thank the Alexander von Humboldt Stiftung for supporting our research.}}
%
%

\author{Frederik Wangelik\orcid{0000-0001-6320-2302}\textsuperscript{\href{mailto:frederik.wangelik@rwth-aachen.de}{\faEnvelopeO}}\and
	Majid Rafiei\orcid{0000-0001-7161-6927}\and
    Mahsa Pourbafrani\orcid{0000-0002-7883-1627}\and
	Wil M.P. van der Aalst\orcid{0000-0002-0955-6940}}
\authorrunning{F. Wangelik et al.}
%
\institute{Chair of Process and Data Science, RWTH Aachen University, Aachen, Germany \\
}
\maketitle              
\begin{abstract}
In recent years, the industry has been witnessing an extended usage of process mining and automated event data analysis. Consequently, there is a rising significance in addressing privacy apprehensions related to the inclusion of sensitive and private information within event data utilized by process mining algorithms.
State-of-the-art research mainly focuses on providing quantifiable privacy guarantees, e.g., via differential privacy, for trace variants that are used by the main process mining techniques, e.g., process discovery.
However, privacy preservation techniques designed for the release of trace variants are still insufficient to meet all the demands of industry-scale utilization.
Moreover, ensuring privacy guarantees in situations characterized by a high occurrence of infrequent trace variants remains a challenging endeavor.
In this paper, we introduce two novel approaches for releasing differentially private trace variants based on trained generative models. With TraVaG, we leverage \textit{Generative Adversarial Networks} (GANs) to sample from a privatized implicit variant distribution. Our second method employs \textit{Denoising Diffusion Probabilistic Models} that reconstruct artificial trace variants from noise via trained Markov chains.
Both methods offer industry-scale benefits and elevate the degree of privacy assurances, particularly in scenarios featuring a substantial prevalence of infrequent variants.
Also, they overcome the shortcomings of conventional privacy preservation techniques, such as bounding the length of variants and introducing fake variants.
Experimental results on real-life event data demonstrate that our approaches surpass state-of-the-art techniques in terms of privacy guarantees and utility preservation.

\end{abstract}
\section{Introduction}\label{sec:intro}

Process mining encompasses a set of data-driven techniques used for the discovery, analysis, and enhancement of business processes. These techniques rely on event data, readily available in various information systems such as ERP, SCM, and CRM systems. Over the past decade, process mining and event data analysis have been effectively implemented across various industries, playing a pivotal role in business success.

Much like other data-driven approaches within the broader field of data science, concerns surrounding the privacy of individuals whose data is subject to process mining algorithms have emerged due to the increasing volume of event data and its utilization. Prominent examples are business process management applications in the health care or governmental sector that use sensitive, personal data records to provide decision support. Consequently, privacy regulations such as GDPR (General Data Protection Regulation) \cite{gdpr} place restrictions on data storage and processing, thereby driving the development of privacy preservation techniques.


Contemporary techniques for preserving privacy primarily rely on Differential Privacy (DP), a privacy framework that introduces controlled noise into data \cite{differential_privacy}. This choice is driven by DP's notable qualities, such as its capacity to offer rigorous mathematical privacy protection and guard against PSO (predicate-singling-out) attacks \cite{PSO}.
The aim of DP-based approaches is to introduce noise into the released output in order to conceal the involvement of an individual. 
Leading-edge research in the field of process mining, incorporating privacy preservation based on DP, focuses on the release of distributions of trace variants. These distributions form the basis for core process mining techniques, namely, process discovery and conformance checking \cite{processMiningBookWil}.
A trace variant refers to a complete sequence of activities performed by an action or agent. Often, this data contains private information. For instance, in the healthcare context, a trace variant refers to a complete sequence of treatment-related activities performed for a patient that is private information itself and can also be exploited to derive other sensitive information, e.g., the disease of the patient. 
Table~\ref{tbl:trace_variant} shows a sample of a trace variant distribution in the healthcare context.
It's important to note that each trace variant within a distribution is linked to an individual, known as a \textit{case} and no case should have more than one associated trace variant \cite{processMiningBookWil}. Hence, this data excerpt motivates that no unauthorized intermediary should be able to link such activities back to individual patients or groups of patients.


\begin{table}[t]
\centering
\scriptsize
\caption{A simple event log from the healthcare context, including trace variants and their frequencies.}
\label{tbl:trace_variant}
    \begin{tabular}{l|c}
    \hline
    Trace Variant   & Frequency \\ 
    \hline
    $\langle register, visit, blood\text{-}test, visit, release \rangle$                    & 13 \\ 
    $\langle register, blood\text{-}test, visit, release \rangle$                    & 14 \\
    $\langle register, visit, hospitalization, surgery, release \rangle$                                       & 4 \\ 
    $\langle register, visit, blood\text{-}test, blood\text{-}test, release \rangle$ & 3  \\ \hline
    \end{tabular}
    \vspace{-0.5cm}
\end{table}

To implement Differential Privacy (DP) for trace variants, conventional methods, often referred to as \textit{prefix-based} approaches, introduce noise from a Laplacian distribution into the variant distribution derived from an event log, as discussed in references such as \cite{felix_differential} and \cite{pripel}.
These approaches involve the generation of all possible unique variants based on a given set of activities to ensure the original distribution of variants is DP-compliant. However, since the number of potential variants that can be generated from a set of activities is infinite, prefix-based techniques must place constraints on the length of the generated sequences. Furthermore, to narrow down the search space, these methods typically employ a pruning parameter to exclude less frequent prefixes.
This DP generation process comes with significant computational complexity and results in several drawbacks, including (1) \textit{introducing fake variants}, (2) \textit{removing frequent true variants}, and (3) \textit{having limited length for generated variants} \cite{rafiei_travag}. 


Several approaches have been put forth to address the challenges mentioned above, either partially or in their entirety.
One such method, known as SaCoFa, as discussed in \cite{sacofa}, aims to alleviate the first and second drawbacks by leveraging insights into the underlying process semantics from the original event data. However, the paper does not delve into the privacy quantification of the additional queries made to acquire knowledge about the underlying semantics. Moreover, the third drawback persists as SaCoFa itself is a prefix-based approach.
In \cite{mineMe} and a related work called Libra \cite{libra}, which builds upon \cite{mineMe}, trace variants are transformed into a representation using a Deterministic Acyclic Finite State Automata (DAFSA) to circumvent the mentioned issues. Nonetheless, Libra introduces a clipping parameter to filter out infrequent variants. This clipping parameter grows based on the number of unique trace variants and the strength of privacy guarantees. Consequently, depending on the number of unique trace variants and privacy parameters, Libra may even eliminate all variants, resulting in empty outputs.
A recent solution, TraVaS, described in \cite{rafiei_travas_short}, proposes an approach based on differentially private partition selection strategies to tackle the aforementioned challenges. Similar to Libra, TraVaS also requires the removal of infrequent trace variants. However, in TraVaS, the threshold for discarding infrequent variants solely depends on the input privacy parameters and does not increase with the number of unique variants or the size of event data. Nevertheless, for small event data with a high prevalence of unique trace variants, TraVaS might face limitations in providing robust privacy guarantees.

In \cite{rafiei_travag}, we introduced \textit{TraVaG} as a fundamentally new approach that incorporates trained, generative models to create differentially private trace variants from an initial input variant distribution.
The core concept behind TraVaG involves privately learning crucial characteristics of event data from an underlying event log through the utilization of deep Autoencoder- and Generative Adversarial Networks (GANs) \cite{GANs}. The GAN, once trained, empowers the generation of new synthetic anonymized variants that closely align with the statistical properties of the original data. To date of publication, it was the first research that had explored the potential of differentially private deep generative artificial neural networks in the domain of process discovery from event log data.
Besides several conceptual advantages over traditional selection-based methods such as a data-independent, memory-efficient application and runtime, a pretraining option, and frequency threshold independence, we exemplified stronger data utility preservation capabilities on real-life event data, especially in DP parameter regions of small $\delta$ and complex variant distributions.

This paper is an extension of TraVaG \cite{rafiei_travag} that expands the experimental verification and analysis of the TraVaG framework and introduces a new generative model in a differentially private setup for transforming confidential event log data into anonymized trace variant samples. The new anonymization framework is based on \textit{Denoising Diffusion Probabilistic Models} (DDPMs) \cite{DDPM15} that already proved to show the best state-of-the-art generative performance on image data \cite{DDPMVis, DDPMStable}. Instead of directly using deep networks to approximate variant statistics, DDPMs are iterative probabilistic models that leverage two types of Markov Chains to first gradually add noise to their training data and then learn to reverse the perturbation by denoising mechanisms. In this paper, we transfer the concept of DDPMs into a differentially private environment with event data structures and compare the approach as well as its performance with our prior GAN-based TraVaG algorithm. This research not only represents a novel effective privatization scheme for the world of process mining, but it is also the first work that investigates the impact of differentially private DDPMs on mixed-type tabular inputs independent of the process mining realm. In addition, we extend the performance analysis of both algorithms to different data structures and more extended ($\epsilon, \delta$) DP parameter regimes. Whereas in \cite{rafiei_travag}, the focus had been put on rather small, complex trace variant distributions, we now also include a larger and more generic as well as realistic sample event log for the utility comparison between DDPMs and our prior TraVaG approach.

Both approaches have different advantages and use cases. Whereas TraVaG provides fast sampling and large flexibility for complex event logs, differentially private DDPMs allow for faster training, more stable conversion, and less complex architectures.
Generally, TraVaG and DDPM as generative models are deployed without data access. Thus, as long as the statistical characteristics of the original data do not significantly change, one does not need to apply DP directly to the original event data. For industry-scale big event data, this property can considerably improve the computational complexities \cite{DP_iterative}.
Moreover, both methods are based on DP-SGD (Differentially Private - Stochastic Gradient Descent) \cite{DP_SGD} optimization techniques that avoid thresholding on training data or released network outputs. Hence, they can generate infinite and arbitrarily large anonymized synthetic trace variants even if the original variant frequencies are comparably small.
Moreover, our experiments on real-life event logs demonstrate the superior performance of both approaches compared to state-of-the-art techniques in terms of data utility preservation for the same privacy guarantees.

The remainder of this paper is structured as follows. In Section~\ref{sec:related_work}, we provide a summary of related work. Preliminaries and notations are provided in Section~\ref{sec:prem}. In Section~\ref{sec:approach}, we present the details of TraVaG and the DDPM framework. Section~\ref{sec:exp} discusses the experimental results based on real-life event logs. In Section~\ref{sec:disc}, a brief summary of privacy, complexity, and data-related challenges is provided, and Section~\ref{sec:conc} concludes the paper.

\section{Related Work}\label{sec:related_work} 

Privacy-preserving process mining is recently growing in importance. 
Several techniques have been proposed to address privacy issues in process mining.
In this paper, our focus is on the combination of generative models and so-called \textit{noise-based} anonymization techniques that are based on the notion of \textit{differential privacy}.
In the following, we thus provide a summary of relevant work focusing on releasing \textit{differentially private event data} and \textit{generating differentially private synthetic data}.

\subsection{Releasing Differentially Private Event Data}
In \cite{felix_differential}, the authors apply an ($\epsilon, \delta$)-Differential Privacy (DP) mechanism to event logs to safeguard the privacy of \textit{directly-follows relations} and trace variants. This approach combines an ($\epsilon, \delta$)-DP noise generator with an iterative query engine, enabling the anonymous release of trace variants with a predefined upper limit on their length.
In a subsequent work, \cite{sacofa}, SaCoFa is introduced as an extension of \cite{felix_differential}. Its primary objective is to optimize query structures by incorporating underlying semantics.
Another extension of \cite{felix_differential} is PRIPEL, described in \cite{pripel}, where additional event attributes are integrated into the privatized event data.

All the above-mentioned techniques follow the prefix-based approach, which has inherent drawbacks, as discussed in Section~\ref{sec:intro}. To tackle these challenges, the authors of \cite{mineMe} introduce a novel method that converts a trace variant distribution into a Deterministic Acyclic Finite State Automata (DAFSA) representation. This approach aims to retain all original trace variants while minimizing the injection of excessive noise during the anonymization process.
A more recent approach, Libra, as outlined in \cite{libra}, builds upon the concepts presented in \cite{mineMe}. Libra focuses on enhancing utility through subsampling and composing privatized subsamples to release differentially private event data.
Additionally, TraVaS, as detailed in \cite{rafiei_travas_short}, presents a novel approach based on differentially private partition selection strategies to address the challenges mentioned in Section~\ref{sec:intro}. Although this method avoids generating fake variants or sequences limited in length, the principle of partition selection still introduces a threshold for infrequent variants.

\subsection{Generating Differentially Private Synthetic Data}

While Differential Privacy (DP)-based generative Artificial Neural Networks (ANNs) have seen substantial research in various data science and machine learning domains, their application in the context of process mining is relatively unexplored. Therefore, we offer a brief overview of relevant work pertaining to structured tabular data.

In \cite{autoenc2}, the primary focus is on the challenge of generating mixed-type labeled data with a choice of $k$ possible labels. The algorithm, known as DP-SYN, initially divides the dataset into $k$ labeled subsets and subsequently conducts private training of an autoencoder on each partition.
In \cite{autoenc1}, a similar approach is adopted, but instead of an anonymous autoencoder, a \textit{variational autoencoder} (DP-VAE) is employed. DP-VAE assumes that the mapping from real data to a Gaussian distribution can be efficiently learned.

Taking a different direction, \cite{syn3} explores the use of a Wasserstein Generative Adversarial Network (WGAN) to generate differentially private mixed-type synthetic outputs, utilizing a Wasserstein-distance-based loss function.
Building on the concepts introduced in \cite{syn3}, \cite{autogan} combines the principles of WGAN and DP-VAE. It first learns a private data encoding and subsequently generates encoded data. This combined approach was adapted by TraVaG to address the challenges posed by the high dimensionality of event data.

Finally, in \cite{TabDDPM}, the authors describe how to train DDPMs on high-dimensional tabular records by introducing multinomial diffusion models for categorical features. Despite promising sampling performance, the authors, however, did not include DP in their training routines. To the best of our knowledge, our paper thus introduces the first differentially private DDPM experiments on complex tabular data.
In the context of non-private generative models for process mining, research primarily focuses on exploiting ANNs and GANs to predict the next state of processes such as \cite{synpro3}, \cite{synpro4}, and \cite{synpro2}.

\section{Preliminaries}\label{sec:prem}

In this section, we introduce the main concepts and definitions utilized throughout the paper. We start with introducing basic notations and mathematical concepts.
Let $A$ be a set. $\multiset(A)$ is the set of all multisets over $A$. 
Given a multiset $B \in \multiset(A)$ over the elements of a set $A$, $B(a)$ is the frequency of $a \in B$.
Given $B_1$ and $B_2$ as two multisets, $B_1 \uplus B_2$ is the sum over multisets, e.g., $[a^2,b^3] \uplus [b^2,c^2] = [a^2,b^5,c^2]$.
We define a finite sequence over $A$ of length $n$ as $\sigma {=} \langle a_1, a_2,\dots, a_n\rangle$ where $\sigma(i) {=} a_i {\in} A$ for all $i {\in} \{1,2,\dots,n\}$. The set of all finite sequences over $A$ is denoted with $A^*$. 

\subsection{\label{subsec:event}Event Data (Log)}

Process mining techniques employ event data, which typically consist of unique events recorded for each activity execution and are characterized by their attributes, e.g.,  \textit{activity} and \textit{timestamp}. 
In this context, a \textit{trace} represents a single execution of a process, comprising a sequence of events related to the same case (individual) and organized in a specific order based on timestamps. Each event can only belong to one trace, and it cannot be repeated within the same trace.
Our research primarily concentrates on the control-flow aspect of event logs, where only the \textit{activity} attribute of events within a trace is considered. This specific perspective is referred to as a \textit{trace variant}. Therefore, we define a simplified event log as a multiset of trace variants.



\begin{definition}[Simple Event Log]\label{def:simple_el}
A simple event log $L$ is defined as a multiset of trace variants $L \in \multiset(\universeActivity^*)$. $\universeLog$ denotes the universe of simple event logs.
\end{definition}

Note that in a simple event log representing a distribution of trace variants, one case, which refers to an individual, cannot contribute to more than one trace variant. At the same time, one trace variant can belong to several cases.

\subsection{\label{subsec:DP}Differential Privacy (DP)}
The main idea of differential privacy revolves around introducing controlled noise into the original data in a manner that makes it practically impossible for an observer to definitively discern whether the information of a particular individual is contained within the data \cite{differential_privacy}. The amount of noise is governed by two key privacy parameters: $\epsilon$, which quantifies the privacy loss (smaller values indicate stronger privacy), and delta $\delta$, which represents the probability of exceeding the privacy loss bound.
In the context of our study, which focuses on simple event logs or the distribution of trace variants as our sensitive event data, we specify the definition of differential privacy as outlined in Definition~\ref{def:dp}.


\begin{definition}[($\epsilon$,$\delta$)-DP for Event Logs]\label{def:dp}
\small
Let $L_1$ and $L_2$ be two neighboring event logs that differ only in a single entry, i.e., $L_2 {=} L_1 {\uplus} [\sigma]$ for any $\sigma {\in} \universeActivity^*$.
Also, let $\epsilon {\in} \mathbb{R}_{>0}$ and $\delta {\in} \mathbb{R}_{>0}$ be two privacy parameters. 
A randomized mechanism $\mechanism_{\epsilon,\delta}{:} \universeLog {\to} \universeLog$ provides ($\epsilon,\delta$)-DP if for all $S {\subseteq} \multiset(\universeActivity^*)$: 
$Pr[\mechanism_{\epsilon,\delta}(L_1) \in S] \leq e^\epsilon {\times} Pr[\mechanism_{\epsilon,\delta}(L_2) \in S] {+} \delta$. 
\end{definition}


In Definition~\ref{def:dp}, $\epsilon$ as the first privacy parameter, specifies the probability ratio, and $\delta$ as the second privacy parameter allows for a linear violation. In the strict case of $\delta = 0$, $\mechanism$ offers $\epsilon$-DP.
The randomness of respective mechanisms is typically ensured by the noise drawn from a probability distribution that perturbs the original trace variant distribution and results in non-deterministic outputs. 
When privacy parameters are set to smaller values, it results in a greater injection of noise into the mechanism's outputs. This, in turn, reduces the probability of deducing the existence of specific instances from these outputs.

\subsection{Generative Adversarial Networks (GANs)}\label{subsec:gan}
Generative Artificial Networks (GANs) represent a class of artificial neural networks designed to generate data samples, often with a focus on capturing the underlying statistical patterns or structures present in the training data \cite{GANs}. These networks are particularly instrumental in various applications, including image generation, natural language processing, and data synthesis.

The fundamental principle behind GANs is to learn a probabilistic model of the data distribution from a given dataset. To accomplish this goal, the models employ a distinctive adversarial training mechanism, which involves two primary network components: a generator $gen: \mathbb{Z}^m \rightarrow \mathbb{R}^n$ and a discriminator $dis: \mathbb{R}^n \rightarrow \{0,1\}$. The generator is responsible for producing synthetic data samples, while the discriminator evaluates these samples to determine whether they are real (from the training data) or fake (generated by the generator). Through a competitive process, the generator continuously improves its ability to produce increasingly convincing data samples, and the discriminator enhances its capacity to distinguish between real and fake data. This adversarial training process often results in the generator becoming proficient at generating data that is difficult to distinguish from authentic data. In this context, it is seeded with random multivariate Gaussian noise $z \in Z^m$ of user-defined dimension $m$ that is converted to synthetic output by the network.
TraVaG applies a GAN architecture to synthesize event data from noise that is similar to the original input.


\subsection{Autoencoders}\label{subsec:autoencoder}
Autoencoders are a class of ANNs designed for unsupervised learning and data compression \cite{autoenc}. They serve a dual purpose; (1) \textit{encoding} input data into a lower-dimensional representation and (2) \textit{decoding} this representation to reconstruct the original data. 
These networks are instrumental in various fields, including image processing, dimensionality reduction, data denoising, and anomaly detection.

The underlying principle of autoencoders involves two primary components: an encoder $enc: \mathbb{R}^n \rightarrow \mathbb{R}^d$ and a decoder $dec: \mathbb{R}^d \rightarrow \mathbb{R}^n$. The encoder processes some high-dimensional input data $x \in \mathbb{R}^n$ and maps it to a compressed representation, often referred to as a bottleneck or latent space $\mathbb{R}^d$ (typically $d \ll n$). The decoder then takes this compressed representation and attempts to reconstruct the original data. During training, the networks $enc$ and $dec$ aim to minimize the difference between the input data and the reconstructed output, which encourages the autoencoder to capture meaningful features and patterns from the data.
We employ the autoencoder principle at TraVaG to learn a reduced encoding of input event data.


\subsection{Denoising Diffusion Probabilistic Models (DDPMs)} \label{subsec:diffusion}

Inspired by nonequilibrium thermodynamics, DDPMs are a class of generative likelihood-based latent variable models that allow matching hidden data distributions by learning to reverse a gradual, iterative noisifying mechanism \cite{DDPM15, DDPM20}. Both noisifying (diffusion) and denoising operations are represented by a combination of two distinct Markov chains. In the course of this work, we follow the approach introduced in \cite{ImprovedDDPM,TabDDPM} and use multivariate Gaussian noise to represent the processes.
Given a data sample $x_0 \in \mathbb{R}^n$ that follows an unknown distribution $x_0 \sim q(x_0)$, we define latent variables of equal dimensionality $x_1 \dots x_T \in \mathbb{R}^n$ through a so-called Markovian \textit{forward process} that adds Gaussian noise at step $t \in \{1 \dots T\}$ with variance $\beta_t \in (0, 1)$ as follows:
\begin{equation}\label{eq.diff1}
    q(x_t|x_{t-1}) = \mathcal{N}(x_t; \sqrt{1-\beta_t}x_{t-1}, \beta_t \mathbf{I}).
\end{equation}
Here, $q(x_t|x_{t-1})$ denotes the conditional probability density distribution of $x_t$ given $x_{t-1}$, $\mathcal{N}(x_t; \sqrt{1-\beta_t}x_{t-1}, \beta_t \mathbf{I})$ represents the Gaussian distribution of $x_t$ with expectation $\mu = \sqrt{1-\beta_t}x_{t-1}$ and variance $\sigma^2 = \beta_t \mathbf{I}$, where $\mathbf{I}$ is the identity matrix of dimension $n$. Considering the joint distribution over all latents $x_1 \dots x_T$ conditional on the data sample $x_0$ then leads to the product
\begin{equation}\label{eq.diff2}
    q(x_1,\dots,x_T|x_0) = \prod_{t=1}^T q(x_t|x_{t-1}).
\end{equation}
When choosing a sufficiently long forward process, i.e., large $T$ and significant variance schedule $\beta_1 \dots \beta_T$, the diffusion chain converges to an isotropic Gaussian distribution of the last variable $x_T$. 
As noted in \cite{ImprovedDDPM}, Equation (\ref{eq.diff3}) further allows accessing the distribution of an arbitrary forward step $t$ directly conditioned on $x_0$:
\begin{equation}\label{eq.diff3}
    q(x_t|x_0) = \mathcal{N}(x_t;\sqrt{\bar{\alpha_t}}x_0,(1-\bar{\alpha_t})\mathbf{I})
\end{equation}
where $\alpha_t = 1-\beta_t$ and $\bar{\alpha_t} = \prod_{s=1}^T \alpha_s$. With standard normally distributed noise $\epsilon$ the latent random variable $x_t$ is therefore expressed as $x_t = \sqrt{\bar{\alpha_t}} x_0 + \sqrt{1-\bar{\alpha_t}}\epsilon$.
To exploit this diffusion principle for generating artificial data samples that follow the same distribution $q(x_0)$, DDPMs complement the noisifying process with a denoising \textit{reverse} Markov chain that is trained to iteratively remove added noise from transformed latents of the forward process.
Starting at Eq. (\ref{eq.diff1}) and the same perturbation schedule, if all posterior distributions $q(x_{t-1}|x_t)$ for $t \in 1\dots T$ were known, we could directly initialize standard-normally distributed latent representations of $x_T$ due to $q(x_T) \rightarrow \mathcal{N}(x_T;0,\mathbf{I})$ for $T \rightarrow \infty$ and simply reverse the forward process to estimate $q(x_0)$ at $t=1$.
However, since $q(x_{t-1}|x_t)$ depends on the entire unknown data distribution, the true posteriors are also unknown and need to be approximated with the help of training samples.
Following the investigations from \cite{DDPM20, ImprovedDDPM}, the best results for the reverse process were achieved by learning a parameterized estimator $p_{\theta}(x_{t-1}|x_t)$:
\begin{equation}\label{eq.diff4}
p_{\theta}(x_{t-1}|x_t) = \mathcal{N}(x_{t-1};\mu_{\theta}(x_t, t),\Sigma_{\theta}(x_t, t))
\end{equation} 
where mean $\mu_{\theta}(x_t, t)$ and variance $\Sigma_{\theta}(x_t, t)$ are represented by deep ANNs with parameters $\theta$.

\section{Approach}\label{sec:approach}

As outlined in Section~\ref{sec:related_work}, DP-based generative ANNs have undergone extensive investigation beyond the realm of process mining. 
Prominent work has revealed common strategies such as the utilization of variational autoencoder architectures, DDPMs, or the integration of GAN architectures.
When transferring these concepts to event data, a pivotal consideration is the management of their intricate high-dimensional structure, which can pose notable challenges during the training process, especially when we introduce noise-based privacy measures into the optimization routines.
Consequently, we have chosen to pursue two cutting-edge methodologies that have demonstrated exceptional efficacy in managing extensive feature spaces. 

First, we demonstrate the idea of TraVaG \cite{rafiei_travag}, which combines the compression functionality of autoencoders with the flexibility of GANs.
Instead of directly generating new event logs, TraVaG first learns a compressed encoding and then trains a GAN to reproduce data within the encoded latent space. Final datasets are obtained by decoding back the dimension-reduced intermediate format.  
This principle mitigates the complication of GANs when extracting statistical properties from feature-rich data that is limited in size. Particularly, sparse features can be compressed without significant loss of information while generator networks improve their learning performance due to the lower dimension. Moreover, no Gaussian Mixture distribution is enforced on the latent space, as it is the case for typical generative stand-alone autoencoder methods \cite{autoenc1}.

Second, we adopt the principle of DDPMs as their design resulted in leading generative performance on broad tabular data as well as large and complex images while being fast and stable to train \cite{TabDDPM, DDPMVis}. In particular, the self-correction capability due to gradual Markovian perturbation processes allows DDPMs to accurately reconstruct underlying high-dimensional data distributions from training samples. For our work, we train a deep network to directly predict the added noise of the forward process given sample trace variants and corresponding noise scheduling step numbers. The incorporation of differential privacy into the training process then provides enough regularization to guarantee simple architectures with fast convergence.

\subsection{Differentially Private GANs and Autoencoders}\label{subsec:TraVaG}

\begin{figure}[t]
\centering
\includegraphics[width=0.99\textwidth]{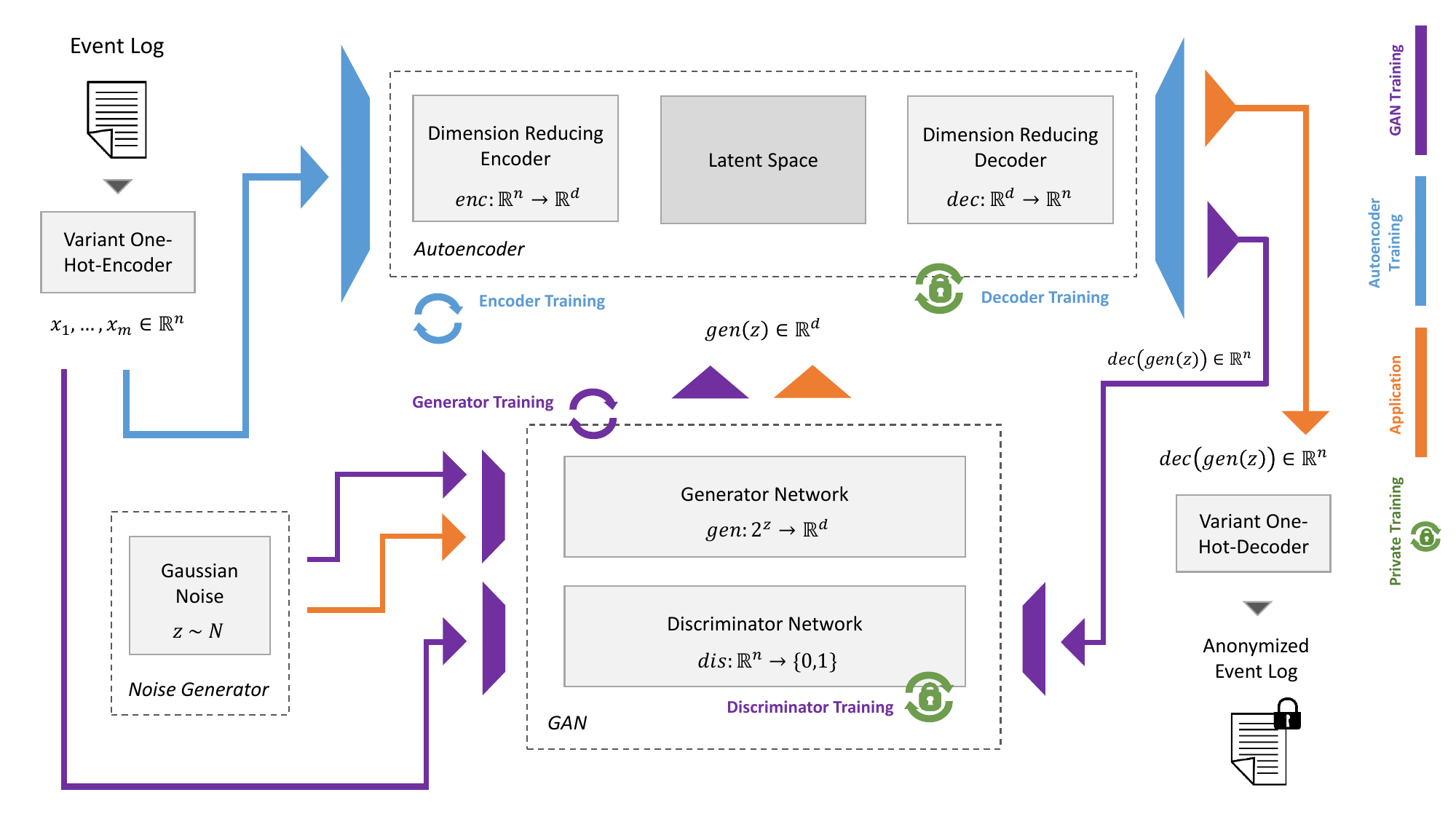}
\caption{Simplified workflow of the TraVaG training and application process \cite{rafiei_travag}.} 
\label{fig:TraVaG_schema}
\end{figure}

The components and workflow of the TraVaG framework are shown in Figure~\ref{fig:TraVaG_schema}.
The process commences with the preprocessing of a basic event log, which contains variant distributions represented as variant-frequency pairs. Here, two common approaches can be taken, each leading to distinct consequences for the final generated outcomes.
The first approach involves examining the activities within variants and extracting all subsequences of immediate neighbors, known as Directly-Follows Relations (DFRs). These DFRs are subsequently transformed into either a binary or numerical space, and are then provided to a GAN as a single feature or as two features, inclusive of their respective frequencies. 
It is essential to note that a drawback of this method is that the generator essentially acts as a sequence constructor, which enables the creation of fake, non-existing variants in the postprocessing phase where all generated activity pairs are interconnected again.

To prevent the generation of such spurious trace variants, we opt for the second alternative, which entails exclusively considering complete variants as input. Thus, a basic event log denoted as $L$ featuring $n$ variants and $m$ cases is encoded in binary form as follows.
In a matrix of dimensions $m \times n$, each variant type corresponds to a binary feature column, while each case corresponds to a row instance. A value of 1 appears in the respective variant column for a given case, with 0s occupying all other positions, thus forming a sparse matrix. Importantly, this transformation can be seamlessly reversed to revert to the original data space after the generation process.
As a result, in contrast to prefix-based methodologies, TraVaG consistently avoids the generation of fake trace variants. Moreover, the one-hot encoding method does not introduce any alterations to the data statistics, thereby incurring no associated privacy costs.
As commonly standardized, we refer to this preprocessing procedure as one-hot encoding and one-hot decoding (see Variant One-Hot-Encoder, Variant One-Hot-Decoder in Figure~\ref{fig:TraVaG_schema}).


Our training process consists of two primary phases: autoencoder training (blue parts) and GAN training (purple parts).
In the following, we provide a broad overview of each training element. Given the central emphasis of our work on privacy considerations, we particularly focus on an in-depth exposition of the privately trained components.
For an exhaustive algorithmic breakdown encompassing the network structures, parameter optimization, activation functions, loss metrics, and optimization techniques, we direct interested readers to consult our supplementary documentation, which is accessible on GitHub.\footnote{\scriptsize \url{https://github.com/wangelik/TraVaGen/blob/main/supplementary/TraVaG_Supplementary.pdf}} 


Following the preprocessing, all sparse binary variant vectors $x_1 \dots x_m \in \mathbb{R}^n$ are first directed to the autoencoder training phase, i.e. to both encoder and decoder ANNs.
These components serve the purpose of converting the high-dimensional data $x_i \in \mathbb{R}^n$ into a more compact representation, the latent space ($\mathbb{R}^d$, $d \ll n$), and vice versa. It is important to note that the dimension $d$ is a hyperparameter of the autoencoder, and its selection is contingent upon the configuration of the GAN. Also, the encoder and decoder are trained differently.
As the encoder does not influence the GAN training process or the generation of new event data, there is no need for it to undergo private optimization, as discussed in \cite{autodp1}, \cite{autoenc2}, and \cite{autoenc1}.
Conversely, the decoder plays a significant role in the anonymization process and is made available to the public. Hence, the corresponding training is carried out with privacy preservation through the utilization of Differentially Private Stochastic Gradient Descent (DP-SGD). Further insights w.r.t. DP-SGD can be found in Section~\ref{subsec:dp_sgd}.

In the next step, the same one-hot encoded data $x_1 \dots x_m \in \mathbb{R}^n$ are used to train a GAN consisting of two feed-forward ANNs; a generator $gen: 2^{\mathbb{Z}} \rightarrow \mathbb{R}^d$ and a discriminator $dis: \mathbb{R}^n \rightarrow \{0,1\}$.
It is important to highlight that the primary objective of the generator, denoted as $gen$, is to generate synthetic data within the output space $\mathbb{R}^d$, which closely resembles the compressed variants. To achieve this, it is initialized with a random multivariate Gaussian noise vector $z$ of user-defined dimension.
On the other hand, the discriminator, labeled as $dis$, is tasked with distinguishing whether its input comes from the decompressed output of the generator $dec(gen(.))$ (classified as fake and assigned to 0), or if it originates from the original data source $x_i, i = 1 \dots m$ (categorized as real and assigned to 1).

Both generator and discriminator components are defined by their network weights and are subjected to an iterative training process in which they engage in a competitive dynamic. The generator's goal is to produce latent space outputs that closely resemble real encoded data, making it challenging for the discriminator to distinguish between fake and real data. On the other hand, the discriminator endeavors to reveal synthetic data records.
Over time, this competitive interplay enables the generator to acquire an understanding of the data and capture the statistical characteristics of the input variant distribution through the perspective of the autoencoder.
It is important to emphasize that due to the integrated autoencoder, the generator exclusively focuses on the latent space $\mathbb{R}^d$, which is notably easier to model in comparison to the intricate data space $\mathbb{R}^n$. Furthermore, this approach effectively prevents the generator from accessing the actual confidential data space. As a result, it does not require training with DP measures, in contrast to the discriminator, which is again privately optimized using DP-SGD algorithms \cite{autogan}.

Finally, after completing the training of both the autoencoder and GAN, TraVaG is ready to be employed for the generation of novel synthetic anonymized event data (orange parts).
The fundamental sampling mechanism mirrors the training phase of the generator. It commences with the generation of a random Gaussian noise sample $z$. This noise is then processed by the generator, producing the output $gen(z)$. From the latent space, the decoder maps this output to $dec(gen(z))$ within the binary data space. Ultimately, the synthetic one-hot encoded result is transformed back into the realm of variant representations.
At this stage, one of the compelling advantages of TraVaG becomes evident in the data format it operates on. Given that the feature space already embodies the various variants present in the original data, TraVaG treats these variants as fixed and merely focuses on learning their distribution during the training process. Consequently, when the framework is put into practice, it can consistently reconstruct an anonymized version of this distribution through multiple iterations, all without the need to create new fake variants.


Generally, the greater the number of synthetic data instances generated, the more refined the resulting TraVaG output becomes. In other words, the newly created anonymized variants progressively approach the distribution of the original variants. Note that this process does not lead to convergence with the actual variant frequencies but rather converges to the internal, learned anonymous representation within TraVaG.
Thus, it is advisable to run TraVaG at least as many times as there are cases in the original event log. In situations where smaller privatized datasets are required, the generated output can be down-sampled during postprocessing rounds.

\subsection{Differentially Private Denoising Diffusion Probabilistic Models}
\begin{figure}[t]
\centering
\includegraphics[width=0.99\textwidth]{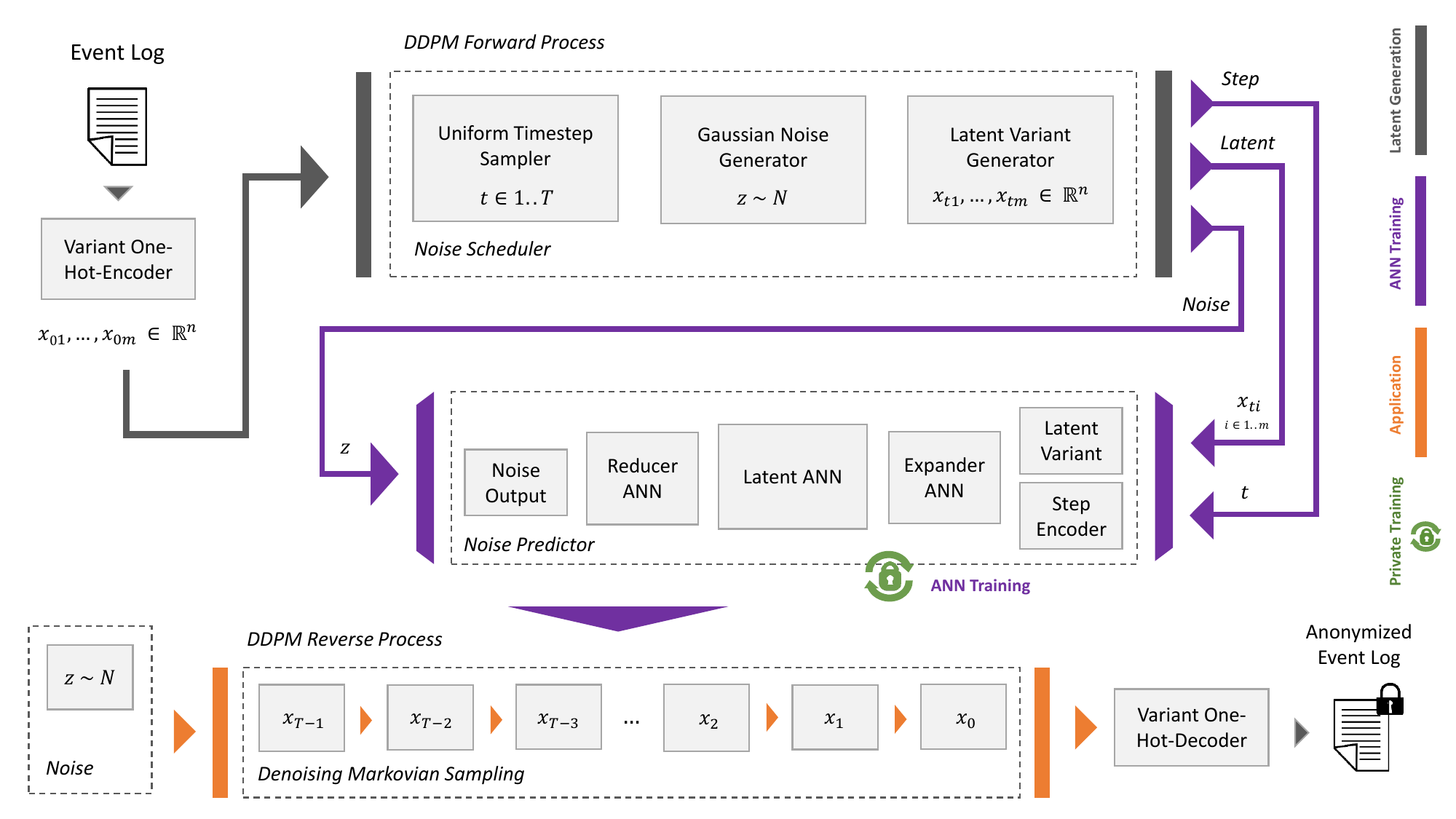}
\caption{Simplified workflow of the DDPM training and application process.}
\label{fig:TraVaDiff_schema}
\end{figure}

The workflow and components of our differentially private DDPM process are shown in Fig. \ref{fig:TraVaDiff_schema}.
Similar to the preprocessing in Subsec. \ref{subsec:TraVaG} all trace variants are initially one-hot-encoded to force the model to only pick up statistical properties of true variants. 
Within the DDPM forward process, we uniformly sample $m$ step numbers from the range $1\dots T$, digest the encoded original variants $x_{01} \dots x_{0m} \in \mathbb{R}^n$ and generate corresponding Gaussian latents $x_{ti} \in \mathbb{R}^n$ for $t \in \{1\dots T\}, i \in \{1\dots m\}$ according to Eq. (\ref{eq.diff3}) (gray components). The outputs of this process are thus triples consisting of step number, latent sample, and added Gaussian noise.

To estimate the posterior according to Eq. (\ref{eq.diff4}) and reverse the diffusion process during sampling, we follow the convention in \cite{DDPM20} that showed remarkable DDPM performance when fixing $\Sigma_\theta(x_t,t)$ to $\bar{\beta}_t = \beta_t (1-\bar{\alpha}_{t-1})/(1-\bar{\alpha}_t)$ and only learning the distribution mean $\mu_\theta(x_t, t)$. Using Eq. (\ref{eq.diff3}) and Bayes theorem, the parameterized mean estimator can be rewritten as
\begin{equation}
    \mu_\theta(x_t, t) = \dfrac{1}{\sqrt{\alpha_t}} \left(x_t - \dfrac{\beta_t}{\sqrt{1-\bar{\alpha}_t}}\epsilon_\theta(x_t,t)\right)
\end{equation}
where $\epsilon_\theta(x_t,t) \in \mathbb{R}^n$ denotes an estimate of the added noise onto $x_0$ at step $t$. Given the training triples of latent variants, true added noise and step number, we train a deep ANN to predict $\epsilon_\theta(x_t, t)$ that has the same dimension as the latent and original encoded variants (purple components). In this process, the step number is first embedded via so-called sinusoidal time embedding blocks\cite{DDPMtime}, concatenated to the latent data, and then forwarded to an expanding ANN layer. Finally, the hidden networks are reduced back to the desired noise dimension. Since the training cycles for predicting $\epsilon_\theta(x_t, t)$ demand repeated access to original variant data, we optimize our ANNs with DP-SGD instead of classical gradient updates similar to the TraVaG framework in Section \ref{subsec:TraVaG}. As a result, all subsequent actions of the DDPM reverse process are differentially private and allow synthetic sampling according to an anonymized internal estimation of the true data distribution.

To eventually start generating new variants from the noise we begin with initializing $x_T$ based on a standard normal distribution $\mathcal{N}(x_T,0,\mathbf{I})$. For all steps $t$ from $T{-}1$ to $1$, the Markovian reverse procedure then iteratively samples denoised latents based on Eq. (\ref{eq:reverse}) until an approximation of $x_0$ is reached \cite{DDPM20}.

\begin{equation}\label{eq:reverse}
    x_{t-1} = \dfrac{1}{\sqrt{\alpha_t}} \left(x_t - \dfrac{\beta_t}{\sqrt{1-\bar{\alpha}_t}}\epsilon_\theta(x_t,t)\right) + \sigma_t z
\end{equation}
Here, $\epsilon_\theta(x_t, t)$ represents the predicted output from our trained DDPM ANN, $\sigma_t = \bar{\beta}_t$ and $z \in \mathbb{R}^n$ denotes a standard normally distributed sample. 
At the end of the sampling routine, all synthetic data are analogously decoded back into their original format. Again, it is recommended to generate at least as many new variants as there used to be in the original dataset to best reconstruct the internally learned estimation of the true data distribution.

\subsection{Differentially Private - Stochastic Gradient Descent (DP-SGD)}
\label{subsec:dp_sgd}
To introduce DP to classical Stochastic Gradient Descent (SGD), Abadi et al. outlined the following two key steps in their work \cite{DP_SGD}.
Given a dataset $X = \{x_i \in \mathbb{R}^n \mid 1 \leq i \leq m\}$, $f$ as a loss function, and $\theta$ as the model parameter. 
First, the gradient $g_i = \nabla_\theta f_\theta(x_i)$ of each data sample $x_i$ is clipped at some real value $C \in \mathbb{R}_{>0}$ to ensure its $L^2$-norm of the gradient does not exceed the clipping value. 
For our work, we make use of the following clipping function\footnote{ \scriptsize Note that also other clipping strategies exist, as highlighted in \cite{DP_iterative}.}:

\begin{equation}
     \text{clip}(g_i,C):=g_i \cdot \min\left(1,C/{||g_i||_2}\right).
\end{equation}

Then, as depicted by Equation~(\ref{eq:noise_gradient_single}), multivariate Gaussian noise parameterized by a noise multiplier $\Phi \in \mathbb{R}$ is added to the clipped gradient vectors before averaging over the batch $B \subseteq X$. Note that we denote the identity matrix as $\mathbf{I}$ and the Gaussian distribution of unspecified dimension as $\mathcal{N}$.

\begin{equation}\label{eq:noise_gradient_single}
\small
g_B \leftarrow\textstyle\frac{1}{|B|}\left(\sum_{i\in B}\text{clip}(\nabla_\theta f_\theta(x_i),C) 
+\mathcal{N}(0,C^2\Phi^2\mathbf{I})\right)
\end{equation}

The noisyfied, clipped and averaged gradient $g_B$ is now differentially private and can be used for conventional descent steps: $\theta \leftarrow \theta - \eta \cdot g_B$, where $\eta$ is the so-called learning rate.
Note that clipping the single gradients as in Equation~(\ref{eq:noise_gradient_single}) can also be replaced by instead clipping gradients of groups of more data points, so-called \emph{microbatches} \cite{DP_iterative}. Here, the initial batch $B$ is, therefore, further partitioned into new batches $B_1, \dots, B_k \subseteq B$ each of size $r$ (skipping the dividend). We then obtain the new microbatch-related gradient, as shown in Equation~(\ref{eq:noise_gradient_mult}).

\begin{equation}\label{eq:noise_gradient_mult}
\small
g_B \leftarrow\frac {1}{k}\left(\textstyle\sum_{i=1 }^k \text{clip}(\nabla_\theta f_\theta(X_{B_i}) 
,C)+\mathcal{N}(0,C^2\Phi^2\mathbf{I})\right).  
\end{equation}

Naturally, standard differentially private SGD (DP-SGD) corresponds to setting $r=1$. 
Increasing the value of $r$ while keeping the size of the batch $|B|$ constant primarily results in decreased training runtime and a reduction in the attained training accuracy. Furthermore, it has been demonstrated to not have a substantial impact on privacy, especially for large datasets \cite{DP_iterative}.

In contrast to the conventional DP parameters $\epsilon$ and $\delta$, DP-SGD employs the noise multiplier $\Phi$ as a control parameter. When transitioning between these two settings, recent research has unveiled a more stringent privacy bound when the batch sampling process for $B$ adheres to a particular Poisson schedule \cite{DP_SGD}. This procedure individually selects each data point from the dataset $X$ with a constant probability denoted as $q$, often referred to as the sampling rate.


\subsection{Privacy Accounting}
\label{subsec:privacy_acc}

To assess and track the precise level of privacy offered by DP-SGD algorithms, we utilize a concept known as \emph{Renyi Differential Privacy} (RDP) \cite{rdp}. RDP represents a distinct notion of differential privacy, primarily employed in the context of private optimization. The underlying mathematical principle is the so-called \emph{Renyi divergence}.
 Given two probability distributions $P$ and $Q$, the Renyi divergence of order $\alpha$ is defined as follows.
 
 \begin{equation}
      D_\alpha(P||Q) := \frac{1}{\alpha-1} \log \mathbb{E}_{x\sim Q} \left( \frac{P(x)}{Q(x)} \right)^\alpha
 \end{equation}




\begin{definition}[($\alpha, \epsilon$)-RDP for Event Logs]
Let $L_1$ and $L_2$ be two neighboring event logs that differ only in a single entry, e.g., $L_2 {=} L_1 {\uplus} [\sigma]$ for any $\sigma {\in} \universeActivity^*$.
Given $\alpha > 1$ and $\epsilon \in \mathbb{R}_{>0}$,
a randomized mechanism $\mechanism_{\alpha,\epsilon}{:} \universeLog {\to} \universeLog$ provides $(\alpha,\epsilon)$-RDP if $D_\alpha(\mathcal{M}(L)||\mathcal{M}(L')) \leq \epsilon$.
\end{definition}

Considering two RDP mechanisms $\mathcal{M}_1$ and $\mathcal{M}_2$, we further denote a composition principle as follows \cite{rdp}.

\begin{proposition}[Composition of RDP]
If $\mathcal{M}_1$ and $\mathcal{M}_2$ are two $(\alpha, \epsilon_1)$-RDP and $(\alpha,\epsilon_2)$-RDP mechanisms for $\alpha > 1$, respectively. Then, the composition of $\mathcal{M}_1$ and $\mathcal{M}_2$ satisfies $(\alpha,\epsilon_1 +\epsilon_2)$-RDP.\label{prop:rdpcomp}
\end{proposition}

Due to the conceptual similarity between $(\alpha, \epsilon)$-RDP and ($\epsilon, \delta$)-DP, the corresponding privacy parameters can be converted \cite{rdp}.

\begin{proposition}[RDP Parameter Conversion]
If a mechanism $\mathcal{M}$ satisfies $(\alpha, \epsilon)$-RDP with $\alpha > 1$, then for all $\delta>0$, $\mathcal{M}$ also satisfies $(\epsilon+(\log 1/\delta)/(\alpha-1), \delta)$-DP.\label{prop:rdpconv}
\end{proposition}

An advantage of using the concept of Renyi divergence during an iterative execution of Gaussian mechanisms, such as for DP-SGD, is that it provides a tighter bound on the privacy loss than standard ($\epsilon,\delta$)-DP composition.
To calculate the final $(\epsilon,\delta)$-DP parameters from multiple runs of RDP-based DP-SGD, a sequence of three steps is required.
Contingent on the chosen sampling strategy, first, a so-called \emph{subsampled Renyi divergence} needs to be derived. Subsequently, privacy levels are aggregated within the framework of RDP before being transformed back into conventional DP.

\begin{enumerate}
    \small
    \item \textbf{Subsampled Renyi Divergence.} Given a sampling rate $q$ and noise multiplier $\Phi$, the RDP privacy parameters for one iteration of DP-SGD can be derived as a non-explicit integral function of $\alpha \geq 1$ \cite{rdp}. This function is standardized in many privacy-related optimization packages and will be referred to as RDP$_1(q,\Phi)$ \cite{DP_SGD}.
    \item \textbf{RDP Composition.} Since DP-SGD is most likely to run iteratively, we need to compose Step 1 over all executions according to Proposition~\ref{prop:rdpcomp}. Hence, the resulting RDP parameters of $T$ iterations are obtained by computing RDP$_T(q,\Phi,T) := \text{RDP}_1(q,\Phi) \cdot T$.
    \item \textbf{Conversion to $(\epsilon,\delta)$-DP.} After retrieving an expression for the overall RDP privacy parameters with RDP$_T$, we need to convert the respective $(\alpha,\epsilon)$ tuple to a $(\epsilon,\delta)$ guarantee according to Proposition~\ref{prop:rdpconv}.
    Note that since the $\epsilon$ parameter of RDP is also a function of $\alpha$, Step 3 involves optimizing for $\alpha$ to achieve a minimal $\epsilon$ and $\delta$.
\end{enumerate}

In the context of our privatized DDPM and TraVaG training algorithms (see Section~\ref{sec:approach}) we employ this accounting procedure to obtain the respective $(\epsilon,\delta)$-DP guarantees for both DDPM ANNs and autoencoder as well as GAN-based discriminator. In the case of TraVaG, the resulting values are then combined into a final privacy cost by the composition theorem of DP \cite{differential_privacy} (see Theorem~\ref{theorem:dp_comp}).

\begin{theorem}[$(\epsilon, \delta)$-DP Mechanism Composition for Event Logs]\label{theorem:dp_comp}
Let $L$ be a simple log, and $\mathcal{M}_i,\allowbreak 1 {\leq} i {\leq} n$ be $(\epsilon_i, \delta_i)$-DP mechanisms. The sequential application of these mechanisms on arbitrary sublogs of $L$ leads to an overall worst-case privacy level parameterized by $(\sum_{1\leq i \leq n} \epsilon_i, \allowbreak \sum_{1\leq i \leq n} \delta_i)$. 
If each $\mathcal{M}_i$ operates on strictly disjoint sublogs of $L$, the worst-case privacy level is $(max_{1\leq i \leq n}~\epsilon_i, max_{1\leq i \leq n}~\delta_i)$, so-called parallel composition.
\end{theorem}

As Theorem~\ref{theorem:dp_comp} states, different $(\epsilon, \delta)$-DP mechanisms can be easily combined into more complex algorithms at the cost of a directly measurable cumulative privacy loss. Nevertheless, the result still promises $(\epsilon, \delta)$-DP independent of the exact form of composition or query structure.

\section{Experiments}\label{sec:exp}

Our experimental evaluation encompasses a broad spectrum of the key privacy parameters, $\epsilon {\in} \{0.001, 0.01, 0.1, 1, 2\}$ and $\delta {\in} \{10^{-6},10^{-5},10^{-4},10^{-3},0.01\}$.
These parameter ranges have been chosen in alignment with typical values utilized in industrial applications and in accordance with contemporary research in the field of DP \cite{felix_differential, sacofa, libra, apple}.
It is worth emphasizing that we have deliberately included extreme settings, such as $\epsilon=2$ and $\delta=0.01$, not because they are practically relevant, but to showcase how the anonymization methods perform when initiated from a weak or non-private baseline.\footnote{\scriptsize Generally, $\delta$ is recommended to be not larger than $1/|D|$, where $|D|$ is the size of dataset $D$ \cite{differential_privacy}.}  

Given the inherent probabilistic nature of $(\epsilon, \delta)$-DP, we execute the TraVaG and DDPM generators 100 times across all input event logs and privacy parameter combinations. Subsequently, we report the average results, as the remaining training-induced standard deviation is small enough to validate all systematic trends. A more detailed study of the generator variance is uploaded to GitHub\footnote{\scriptsize \url{https://github.com/wangelik/TraVaGen/blob/main/supplementary/Uncertainty_Supplementary.pdf}}.
For comparison, we assess our findings against TraVaS, an established state-of-the-art technique \cite{rafiei_travas_short}, and the original prefix-based framework, denoted as the \emph{benchmark} \cite{felix_differential}.
Note that in \cite{rafiei_travas_short}, TraVaS was already compared against SaCoFa \cite{sacofa} and the benchmark approach from \cite{felix_differential}, and exhibited superior performance. In this work, we have included the benchmark method to facilitate straightforward comparisons.
Additionally, it is important to mention that Libra \cite{libra} does not accept $\epsilon$ as an input parameter but instead computes it based on $\alpha$ as an RDP parameter and the applied sampling strategy. This aspect complicates direct comparisons based on exact $\epsilon$ and $\delta$ parameters. Nevertheless, one notable observation, in contrast to our generative models, is that Libra returns an empty log for event datasets with numerous infrequent variants (such as Sepsis) when $\delta \leq 10^{-3}$.

The configuration of the ANNs in our generative models is based on a semi-automated tuning approach tailored to the specific input logs.
While many design choices and hyperparameters are adjusted based on results from manual testing and research experience, certain parameters, including the \emph{batch size} ($B$), the \emph{number of iterations} ($I$), and the \emph{noise multiplier} ($\Phi$), are automatically optimized using a grid-search methodology for fixed privacy levels \cite{gridsearch}.
A comprehensive list of all resulting configurations for each event log can be found on GitHub.\footnote{\scriptsize \url{https://github.com/wangelik/TraVaGen/tree/main/supplementary}}

\subsection{Datasets}\label{subsec:datasets}

We investigate the algorithm performance using real-life event data. For this purpose, three event logs with varying sizes and levels of trace uniqueness have been selected. As previously discussed in Section~\ref{sec:intro} and highlighted in other research papers such as \cite{felix_differential}, \cite{sacofa}, \cite{rafiei_group}, and \cite{libra}, privatizing infrequent variants can be particularly challenging. Hence, trace uniqueness serves as a crucial metric for our analysis.
The first dataset, known as the Sepsis log, documents hospital processes for Sepsis patients and is notable for containing numerous rare traces \cite{sepsis}.
In contrast, the BPIC-2013 dataset encompasses a significantly larger number of cases but also exhibits a trace uniqueness that is four times smaller compared to the Sepsis log. BPIC-2013 pertains to an incident and problem management system known as VINST \cite{bpic13}.
Lastly, the BPIC-2012-App log from \cite{bpic12} reports process data associated with various loan applications from a Dutch financial institution. This dataset offers lower-dimensional entries with relatively small trace uniqueness.
Note that in the experimental verification, our focus is on data with diverging variant distributions and not primarily large sizes. As a result, big, yet, at the same time, similar-in-shape event logs such as the \textit{Road Traffic Management} dataset \cite{road_traffic} are not considered. With 150370 traces over 231 variants, \textit{Road Traffic Management} approximately represents a scaled version of the BPIC-2012-App log, and its analysis would provide insights into the resource-dependent training time rather than on the algorithm's capabilities of picking up complex frequency distributions.
Moreover, it is important to underline that all of these logs are authentic examples of confidential human-centric data where the case identifiers are linked to individuals.
For more detailed log statistics, we refer to Table~\ref{tab:exp_data}.

\begin{table}[tb]
\scriptsize
\centering
\caption{General statistics of the event logs used in our experiments.}
\label{tab:exp_data}
\begin{tabular}{c|c|c|c|c|c}
\hline
Event Log & \#Events & \#Cases & \#Activities & \#Variants & Trace Uniqueness\\
\hline
Sepsis & 15214 & 1050 & 16 & 846 & 80\%\\
BPIC-2013 & 65533 & 7554 & 4 & 1511 & 20\%\\
BPIC-2012-App & 60849 & 13087 & 10 & 17 & 0.12\%\\
\hline
\end{tabular}
\end{table}

\subsection{Evaluation Measures}\label{subsec:eval_measures}
To assess the effectiveness of a $(\epsilon, \delta)$-Differential Privacy (DP) mechanism in preserving the utility of data or results, it is crucial to utilize suitable evaluation metrics.
The perspective of \textit{data utility} involves assessing the resemblance between two logs, irrespective of their potential future applications. To compute data utility, we rely on specific measures, including \textit{relative log similarity} \cite{rafiei_quant,rafiei_travas_short} and \textit{absolute log difference} \cite{rafiei_travas_short,rafiei_travag}.


The \textit{relative log similarity} metric assesses the \textit{earth mover's distance} between two distributions of trace variants. It employs the normalized Levenshtein string edit distance as the similarity function to measure the resemblance between trace variants. Consequently, this metric quantifies how closely the variant distribution in an anonymized log aligns with the original variant distribution, with values ranging from 0 to 1.


\textit{Absolute log difference} accounts for the situations where distribution-based measures provide misleading expressiveness \cite{rafiei_travas_short}.  
An instance of this is when event logs exhibit similar variant distributions but significantly differ in size.
To calculate the \textit{absolute log difference} value, we adopt the methodology introduced in \cite{rafiei_travas_short}. First, it involves transforming the input logs into a \textit{bipartite graph} where variants are treated as vertices. Subsequently, a \textit{cost network flow} problem is solved, with demands and supplies being determined based on the absolute variant frequencies. In this context, the associated edge costs are determined by the absolute Levenshtein distance between variants.
As a result, the outcome of this optimization signifies the minimum number of Levenshtein operations necessary to convert variants in an anonymized log into variants found in the original log.
More detailed documentation on the exact algorithms is provided on GitHub.\footnote{\scriptsize \url{https://github.com/wangelik/TraVaGen/blob/main/supplementary/Metrics_Supplementary.pdf}}


In addition, we analyze the performance of our methods with respect to \textit{result utility preservation} specifically in the context of \textit{process discovery}, which is a specialized application relying on trace variant distributions.
To conduct this assessment, we employ the \textit{inductive miner infrequent} algorithm \cite{sander_inductive}, setting a default noise threshold of 20\% to derive process models from the anonymized event logs for all the investigated $(\epsilon, \delta)$ settings. Then, these resulting models are compared with the original event log to compute token-based replay scores for \textit{fitness} and \textit{precision}, as outlined in \cite{processMiningBookWil}.

\subsection{Data Utility Analysis}\label{subsec:exp_datautil}
\begin{figure}[ht!]
\centering
\includegraphics[height=0.8\textheight, keepaspectratio]{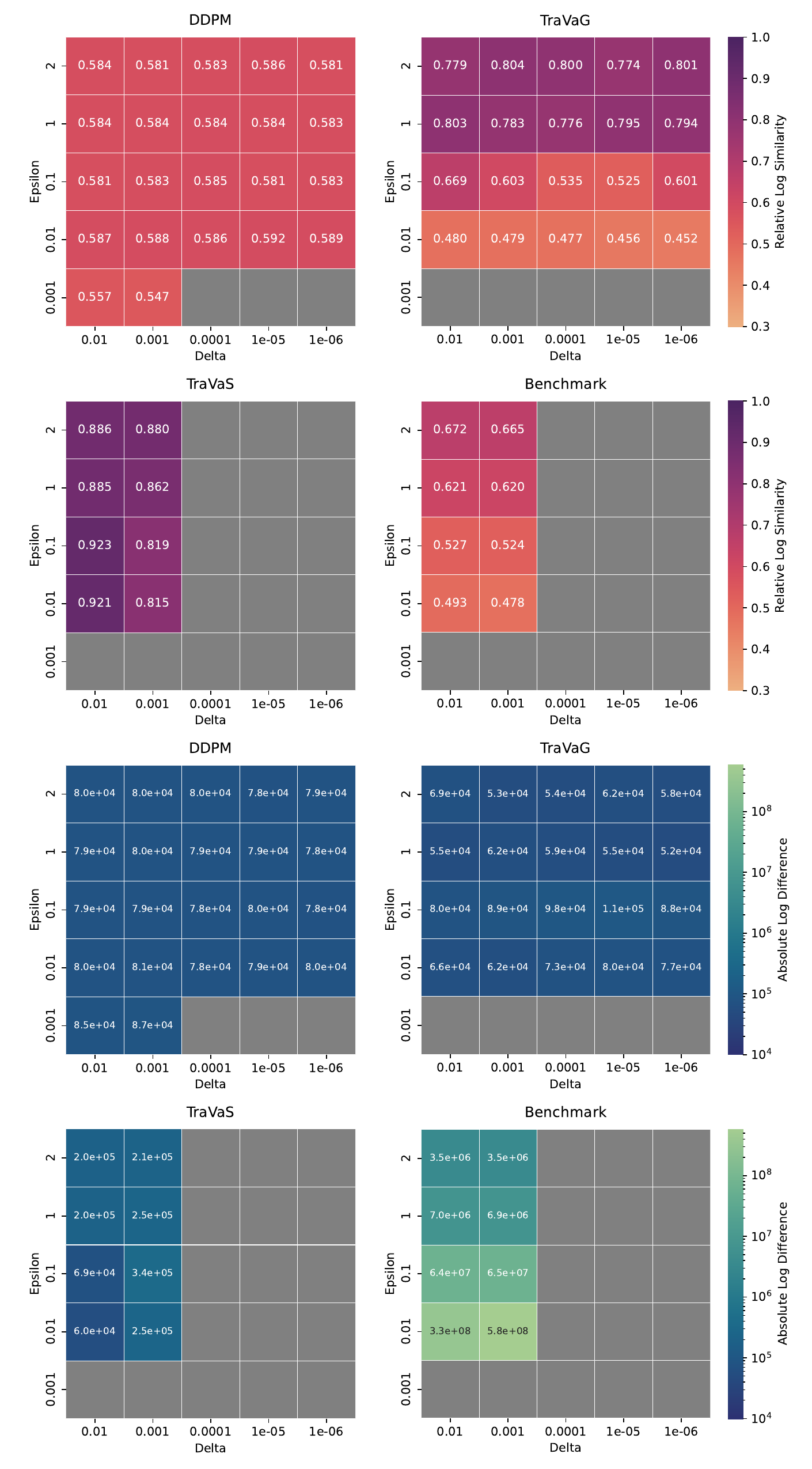}
\caption{The \emph{relative log similarity} and \emph{absolute log difference} results of anonymized BPIC-2013 logs generated by DDPM, TraVaG, TraVaS, and the benchmark. Each value represents the mean of 100 generations for DDPM, TraVaG, and 10 algorithm runs for TraVaS and the benchmark.} \label{fig_exp1}
\vspace{-0.3cm}
\end{figure}

In this subsection, the results of the two aforementioned data utility metrics are presented for all three real-life event logs.
Figure~\ref{fig_exp1} shows the average results on BPIC-2013 in an eight-fold heatmap. The gray fields denote an unsuccessful algorithm execution. For $\delta<10^{-3}$, the thresholding of TraVaS becomes too strict and removes many variants in the anonymized outputs. On the contrary, the benchmark introduces artificial variants and noise to an extent that is unfeasible to average within reasonable time and accuracy.
In opposition, our novel DDPM and TraVaG approaches successfully manage to generate anonymized outputs for $\delta<10^{-3}$. Due to more stable training and less noise introduction, the diffusion principle even allows working in the high $\delta$-regime for $\epsilon = 0.001$. 

\begin{figure}[t!]
\centering
\includegraphics[height=0.8\textheight, keepaspectratio]{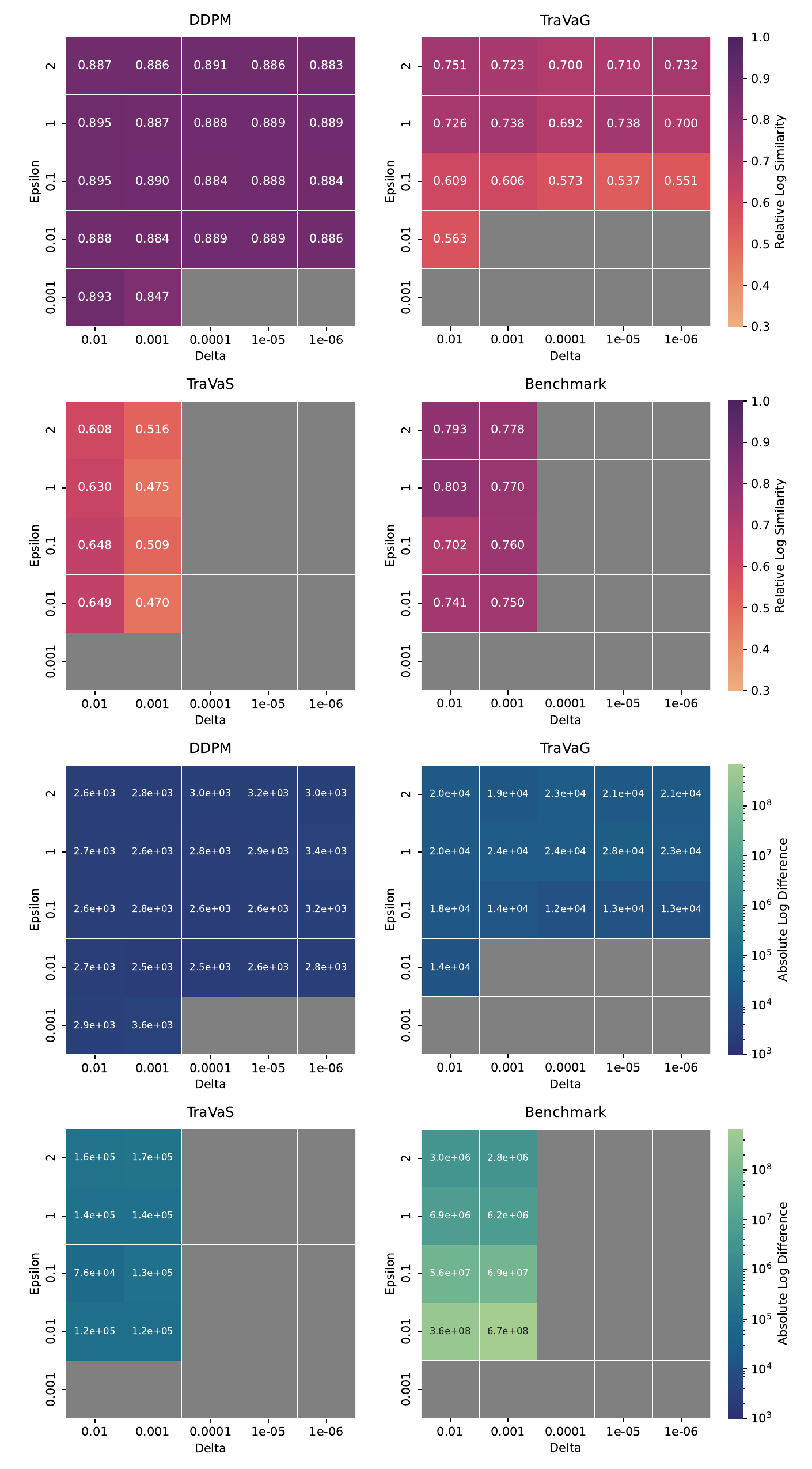}
\caption{The \emph{relative log similarity} and \emph{absolute log difference} results of anonymized Sepsis logs generated by DDPM, TraVaG, TraVaS, and the benchmark. Each value represents the mean of 100 generations for DDPM, TraVaG, and 10 algorithm runs for TraVaS and the benchmark.} \label{fig_exp2}
\vspace{-0.3cm}
\end{figure}

\begin{figure}[ht!]
\centering
\includegraphics[height=0.8\textheight, keepaspectratio]{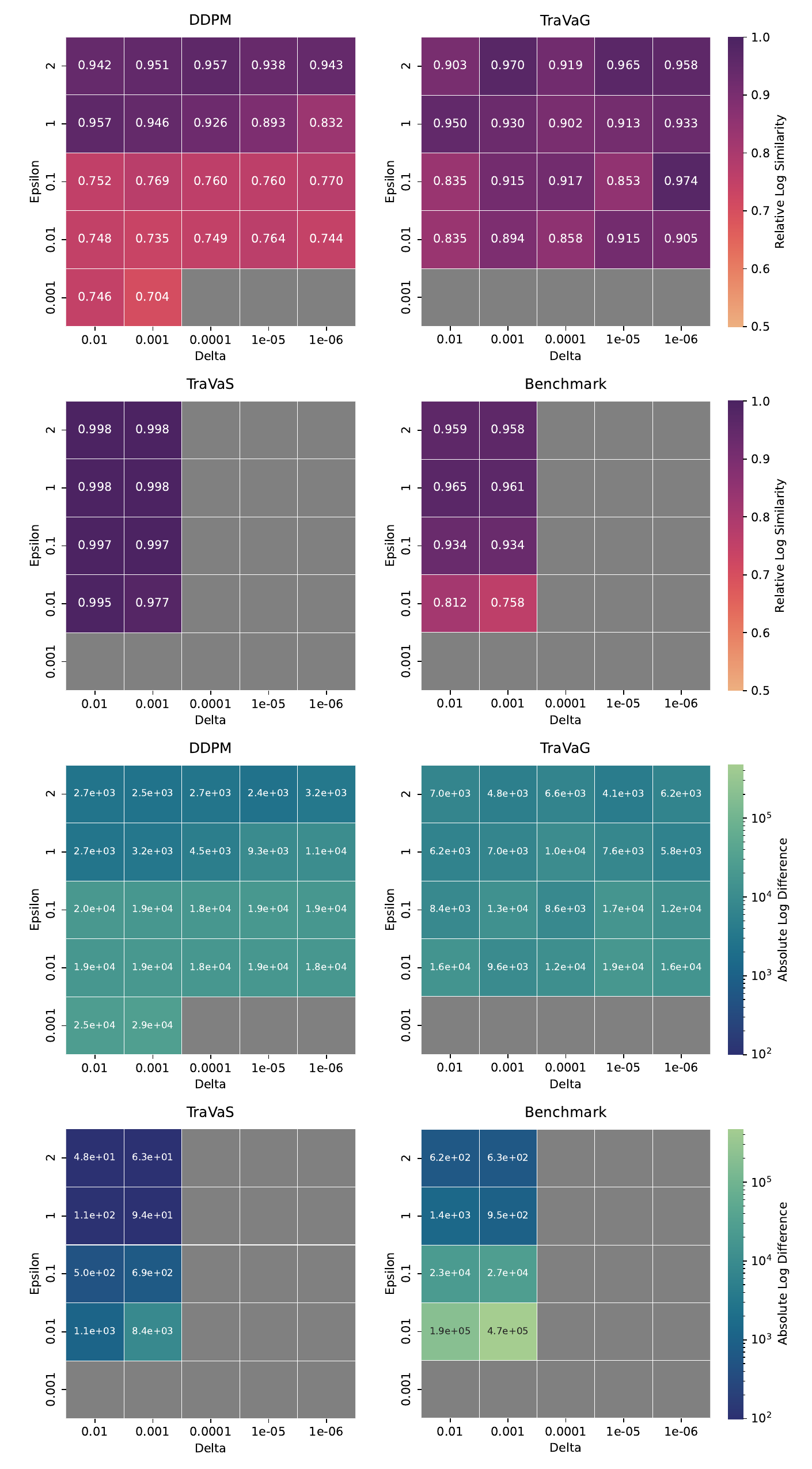}
\caption{The \emph{relative log similarity} and \emph{absolute log difference} results of anonymized BPIC-2012-App logs generated by DDPM, TraVaG, TraVaS, and the benchmark. Each value represents the mean of 100 generations for DDPM, TraVaG, and 10 algorithm runs for TraVaS and the benchmark.} \label{fig_exp3}
\vspace{-0.3cm}
\end{figure}

More importantly, both DDPM and TraVaG results of \emph{relative log similarity} and \emph{absolute log difference} do not illustrate clear decreasing trends on lower $\delta$ within the investigated parameter range. We explain this expected observation by the fact that our trained generative models avoid any pruning mechanism on their output and implement less $\delta$-dependent Gaussian noise via RDP into the gradients (see Section~\ref{subsec:privacy_acc} and \cite{rdp}).
Whereas the absolute log difference results maintain a rather stable output for the different ($\epsilon,\delta$) values, the relative log similarity of TraVaG presents a strong positive $\epsilon$-dependency. As a result, the absolute statistics (absolute Levenshtein distances and absolute frequencies) of the anonymized event data seem to be more similar to the original logs as the variant distributions with increasing noise. A rationale for this discrepancy lies in the comparably small dataset with 7554 instances over 1511 variants. By construction, TraVaG accomplishes reproducing equally sized event logs containing many original variants but fails to pick up some characteristics of the underlying distribution once the input data or the training iterations are limited. Hence, we expect this diverging trend to diminish with increasing training data.
Our results from using DDPM demonstrate consistent performance across both evaluation metrics. Notably, DDPM outperforms TraVaG in terms of \textit{absolute log difference}, although it shows a slightly lower performance in \textit{relative log similarity} in the higher $\epsilon$-regime.
This observation resembles the different use cases and trade-offs provided by the different model architectures. As our DDPM framework comprises more complex ANNs that adopt high-dimensional variant distributions less accurately with limited training data than TraVaG, the model releases lower relative similarity scores for weaker privacy settings.
However, when the privacy guarantees are increased, their iterative processes lead to self-correction of introduced noise that compensates for further performance decrease and keeps the results more invariant. 

The data utility results for the Sepsis log are presented in Figure~\ref{fig_exp2}.
With only 1050 instances at 846 variants, this dataset is even smaller and more trace-unique than BPIC-2013. Nevertheless, the overall dimensionality shows a significant reduction.
As a result, we observe similar, but more pronounced behavior of \textit{relative log similarity} and \textit{absolute log difference} metrics compared to Figure~\ref{fig_exp1}. Extreme examples are the metrics at $\epsilon<0.01, \delta<10^{-2}$, where the introduced gradient noise turned out as too intense for the TraVaG model to converge under the given training data size. 
In contrast, the DDPM that only integrates one privately trained ANN compound and employs gradual noise perturbation, again successfully works even at the low $\epsilon$-regime.
For the remaining privacy settings, both DDPM and TraVaG outperform their competitors with respect to the absolute log statistics, while the relative log similarity shows leading performance for DDPM and similar results for TraVaG compared to TraVaS and the prefix-based benchmark.
The main cause is rooted in the dataset structure where the lower dimensionality is better manageable for the DDPM architecture so that its gradual denoising advantage leads to stable and unmatched log similarities.

Last but not least, Figure~\ref{fig_exp3} depicts the performance evaluation of DDPM, TraVaG, TraVaS, and the benchmark on the BPIC-2012-App event log comprising 13087 samples over 17 unique variants.
In contrast to Sepsis and BPIC-2013, both \textit{relative log similarity} and \textit{absolute log difference} for DDPM and TraVaG now indicate a slightly increasing data quality with increasing $\epsilon$.
Combined with the generally superior metric scores in the less strict privacy regime, this pattern can be understood by a better-learned variant distribution due to the larger training input.
Interestingly, DDPM underperforms our TraVaG model for $\epsilon<1$ and outperforms at $\epsilon>0.1$. Due to the fundamental differences in the training principle, we assume the gradual denoising chain of the DDPM framework to be suboptimal at strong DP in the event of balanced, low-dimensional data distributions compared to the direct sampling ANN of TraVaG.
In addition, despite the log-related performance boost, we notice a considerable underperformance with respect to TraVaS and benchmark at $\delta>10^{-4}$.
Since both TraVaS and the benchmark release variant data by noise addition and thresholding, large event logs with many frequent traces (such as BPIC-2012-App) are hardly affected as long as the threshold is lower than the lowest frequency. On the contrary, our generative models still have to privately learn the underlying data distribution during multiple training iterations, which is more prone to errors and slight deviations.
Nevertheless, we note that the general ability of DDPM and TraVaG to capture relevant data characteristics also significantly increases with more frequent variants (see Figure~\ref{fig_exp3}), particularly if enough training data is available.

\subsection{Process Discovery Analysis}\label{subsec:exp_procdisc}

\begin{figure}[t!]
\centering
\includegraphics[height=0.8\textheight, keepaspectratio]{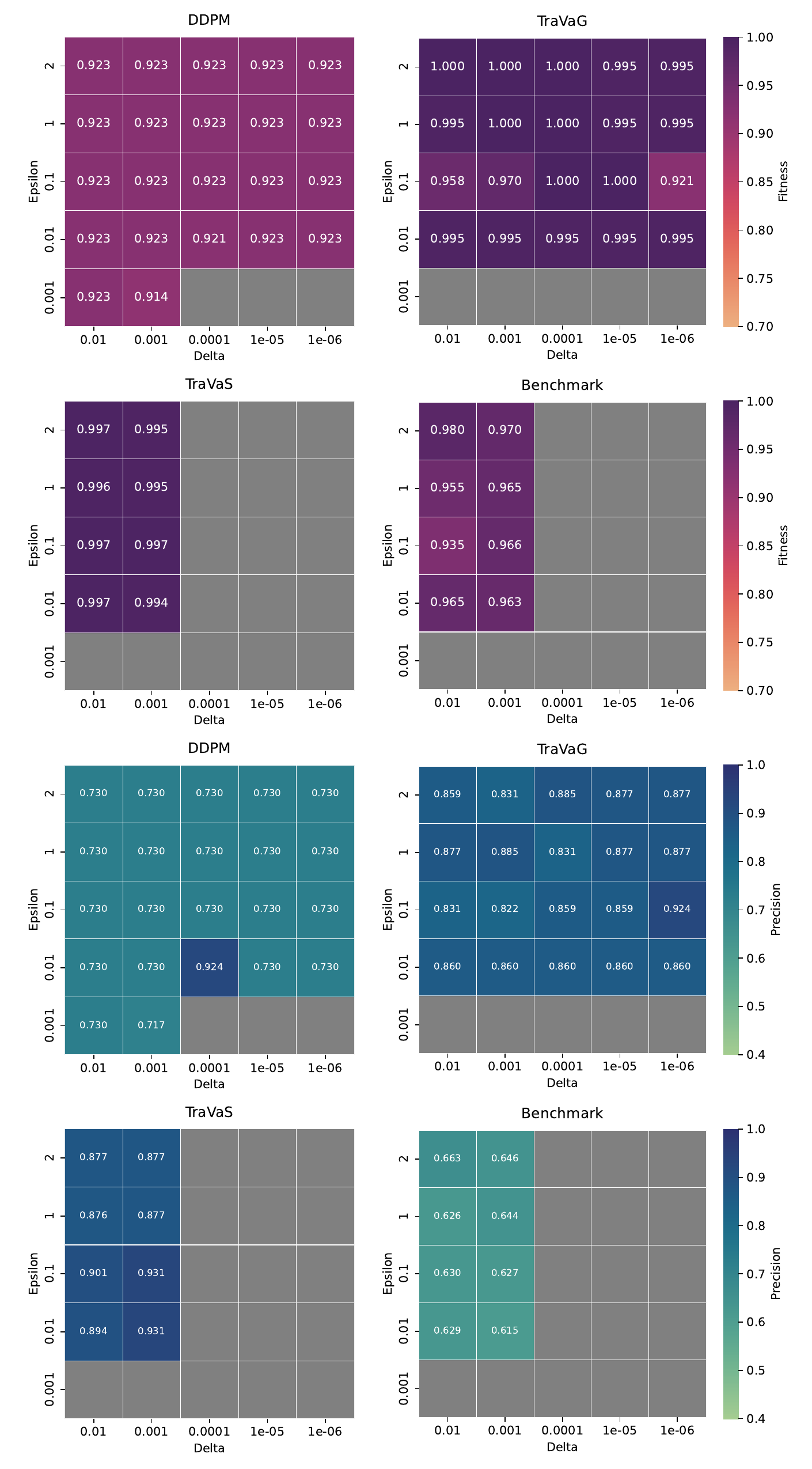}
\caption{The \textit{fitness} and \textit{precision} results of anonymized BPIC-2013 event logs generated using DDPM, TraVaG, TraVaS, and the benchmark. Each value represents the mean of 100 generations for DDPM, TraVaG, and 10 algorithm runs for TraVaS and the benchmark.} \label{fig_exp4}
\vspace{-0.3cm}
\end{figure}

\begin{figure}[t!]
\centering
\includegraphics[height=0.8\textheight, keepaspectratio]{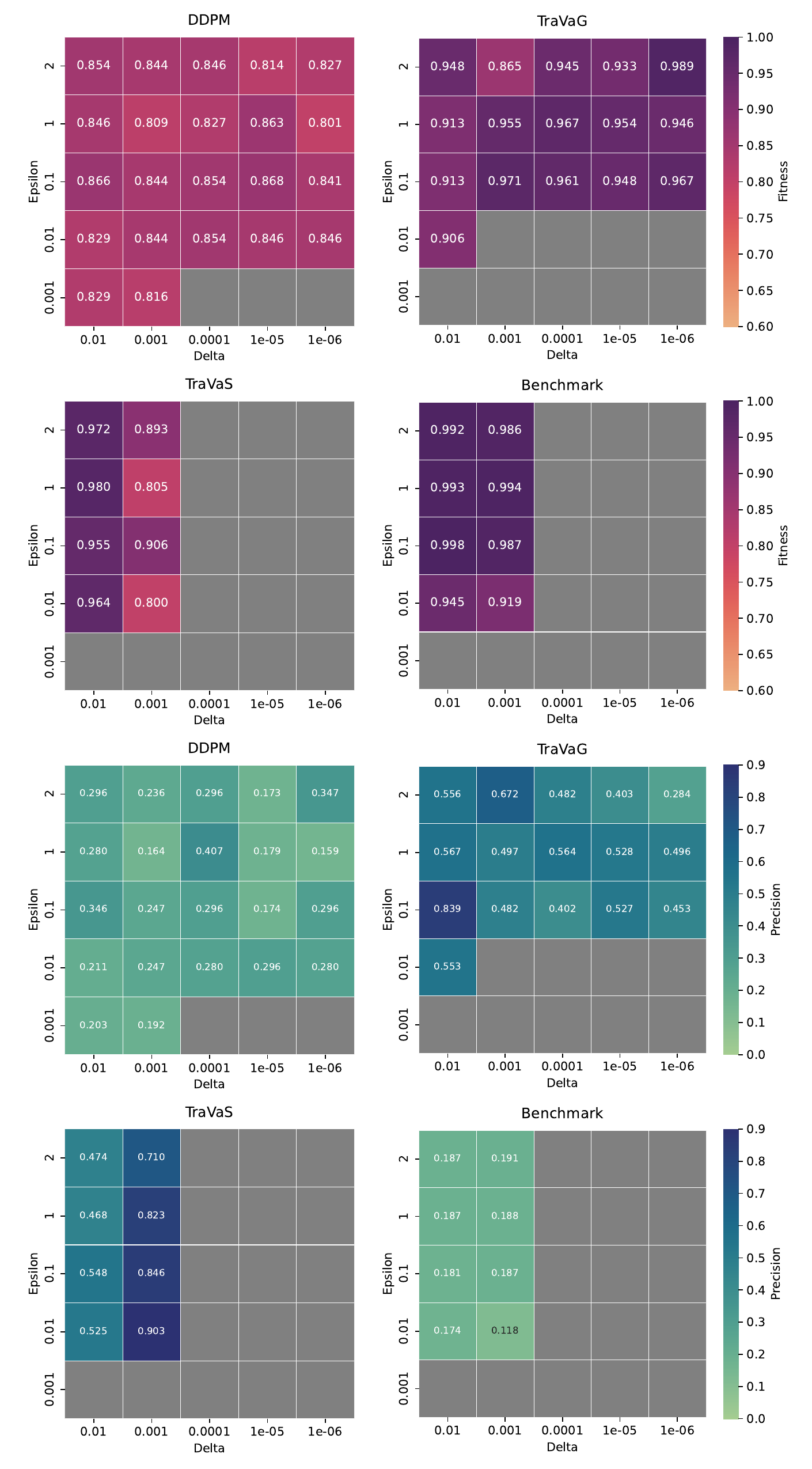}
\caption{The \textit{fitness} and \textit{precision} results of anonymized Sepsis event logs generated using DDPM, TraVaG, TraVaS, and the benchmark. Each value represents the mean of 100 generations for DDPM, TraVaG, and 10 algorithm runs for TraVaS and the benchmark.} \label{fig_exp5}
\vspace{-0.3cm}
\end{figure}

\begin{figure}[ht!]
\centering
\includegraphics[height=0.8\textheight, keepaspectratio]{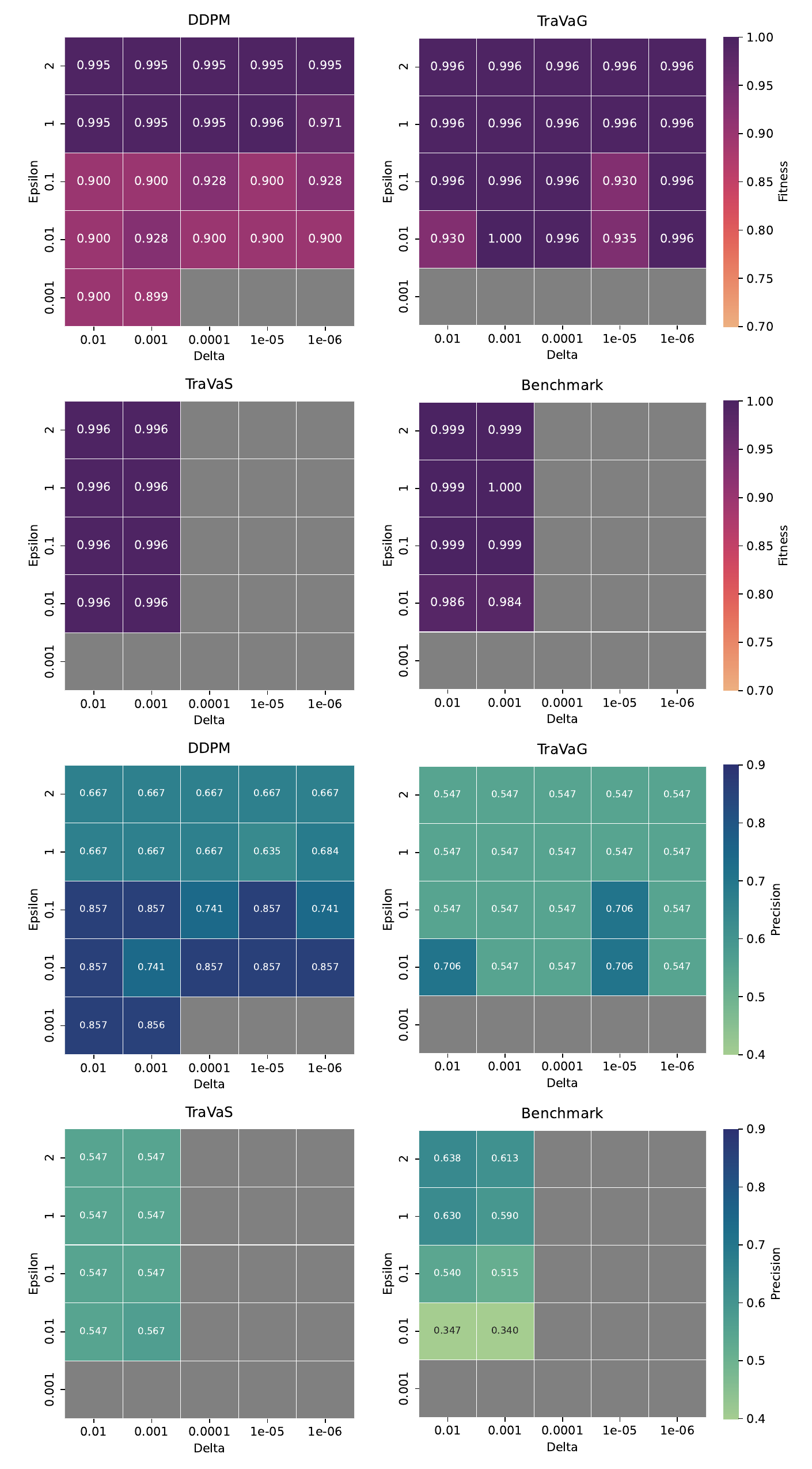}
\caption{The \textit{fitness} and \textit{precision} results of anonymized BPIC-2012-App event logs generated using DDPM, TraVaG, TraVaS, and the benchmark. Each value represents the mean of 100 generations for DDPM, TraVaG, and 10 algorithm runs for TraVaS and the benchmark.} \label{fig_exp6}
\vspace{-0.3cm}
\end{figure}

Our data utility analysis is complemented with a process discovery evaluation based on \emph{fitness} and \emph{precision} scores.
Figure~\ref{fig_exp4} illustrates the result utility analysis of DDPM, TraVaG, TraVaS, and the benchmark on the BPIC-2013 log.
As discussed in Subsection~\ref{subsec:exp_datautil}, both generative models successfully manage to produce results for small $\delta<10^{-3}$ where the other methods are not applicable.
Except for the three outliers at $\epsilon = 0.1$, both fitness and precision show a stable distribution without considerable dependence on the different privacy parameters and with slight outperformance of TraVaG. We thus conclude that the absolute log difference provides a better proxy for the process-discovery-based performance of DDPM and TraVaG models than relative log similarity.
Similarly, the strong scores on both metrics demonstrate a sufficient replay behavior between the model obtained from an anonymized log and the original log. Whereas fitness denotes that the process model still captures most of the real underlying event data, precision depicts only a small fraction of model decisions, not being included in the original event log.
Consequently, TraVaG and DDPM accomplish learning the most important facets of the BPIC-2013 variant distribution for the discovery algorithm to produce a fitted model.

The result utility evaluation of the high trace-unique Sepsis log is presented in Figure~\ref{fig_exp5}.
With respect to fitness, TraVaG shows similar values as TraVaS but a slight under-performance compared to the benchmark. The main cause for this observation again refers to the infrequent variants and the small log size. While TraVaS maintains a strong $\delta$-related threshold and TraVaG copes with the limited training data, the benchmark introduces many artificial variants but tends to match the frequent traces. As a result, the discovered process models are able to replay most of the original behavior, in contrast to TraVaG and TraVaS results.
According to the aforementioned explanation, precision reflects an inverted trend.
Here, the process models resulting from the anonymized event logs using the benchmark technique contain many possible behaviors that are nonexistent in the underlying event log. For TraVaS and TraVaG, we thus achieve more precise anonymized process models.
Interestingly, our DDPM framework shows slightly worse fitness and precision performance than TraVaG, despite being superior in the data utility analysis (see Figure~\ref{fig_exp2}). We explain this trend by the fact that the DDPM generator tends to oversample infrequent features and under-represent distribution peaks with limited training input more than TraVaG. Whereas activity-based utility scores optimize and measure over the entire variant distribution and are therefore rather invariant to slight deviations, noise-thresholded process discovery algorithms can lead to more pronounced effects. These disparities are anticipated to diminish as more training data is employed for DDPM.

Figure~\ref{fig_exp6} demonstrates fitness and precision for all anonymization techniques on the BPIC-2012-App event log.
We first highlight the almost constant values of TraVaG and TraVaS which underline that the different noise levels per $(\epsilon,\delta)$ setting of the privatized outputs do not leak through the noise-thresholded process discovery algorithm.
This observation can be traced back to Figure~\ref{fig_exp3}, where the different data utility metrics still show $(\epsilon,\delta)$ variations as expected. In particular, the slight under-performance of TraVaG compared to TraVaS does not seem to be relevant in the context of anonymized process discovery.
On the contrary, the DDPM results reflect the data utility variation and its explanation from Figure~\ref{fig_exp3}, where the gradual sampling technique turned out to be inferior to our GAN setup at the low $\epsilon$-regime.
Accordingly, the discovered process models show less accurate replay behavior of the original log (lower fitness) as some more infrequent variants were undersampled during the generation. In direct correspondence, the slightly reduced, privatized models lead to fewer unmatched behavior and thus better overall precision.
Finally, we briefly note the three artifacts at $(\epsilon=0.1, \delta=10^{-5})$, $(\epsilon=0.01, \delta=0.01)$ and $(\epsilon=0.01, \delta=10^{-5})$ for TraVaG. Although the resulting scores are still in a similar range as the remaining values, the increased precision at a decreased fitness leads to the assumption of some missing parts within the corresponding privatized models. Such events may occasionally occur if TraVaG fails to learn a specific frequent variant from the underlying data. In fact, the explanation is based on the same assumption that supports the performance variation in the DDPM results.

\section{Discussion}\label{sec:disc}

The subsequent discussion expands upon the findings derived from the analysis conducted in Section \ref{sec:approach} and Section \ref{sec:exp}, focusing on delineating various structural, operational, and technical attributes of the TraVaG and DDPM algorithms in greater depth. First, we discuss the privacy limitations induced by our DP framework. Then, we describe a comparative assessment of complexity from both training and application perspectives. Lastly, elucidation is provided regarding the specification of input data requirements.

We opted for DP as an anonymization technique due to its mathematical privacy notion plus its security, quantifiability, and widespread adoption in industrial contexts. However, alongside these advantages, there exist some drawbacks and limitations. These include the intricate and less intuitive formalism of DP, limited applicability within algorithms, and challenges in interpreting its parameters $(\epsilon, \delta)$ \cite{differential_privacy}. Despite dedicated efforts in research focusing specifically on elucidating and tuning DP, achieving transparency in DP-based data protection for uninformed users remains a persistent challenge.
Furthermore, it is important to highlight that our data format, particularly in terms of variant frequencies, results in DP protecting individuals who contribute to specific variants through their recorded cases instead of the variants themselves. Given the probabilistic nature of DP and the threshold-independent sampling utilized by our generative models, there may arise privacy-critical scenarios that warrant attention and may hold practical significance.
As an illustration, we consider an informed attack model where an adversary possesses access to a trained TraVaG or DDPM generator and is aware that the appearance of a particular variant implies association with cases of a specific individual. Since our generative models retain only true variants without truncating low frequencies, during training, there exists a probability of capturing this variant if it is present in the input event log. In turn, the adversary can exploit upon its occurrence in the sampling process to infer the specific individual. Nevertheless, without detailed domain knowledge, individual cases are still protected by both, the aggregation within variant distributions and the noise insertion of DP, despite the design principle of models exclusively releasing true variants.

When analyzing and comparing model complexity for TraVaS, TraVaG, and DDPM algorithms, the different algorithmic classes of the underlying frameworks have to be considered and differentiated. For this paper, all computations were conducted utilizing one NVIDIA Tesla P100 GPU and an Intel Xeon 2.20 GHz CPU.
As explained in \cite{rafiei_travas_short}, TraVaS operates as a selection-based technique devoid of explicit training requirements. Instead, it dynamically introduces specific noise during runtime into the variant distribution, a process that scales with the number of variants and can be parallelized. Consequently, the application runtime remains under 1 second, albeit necessitating the event log being loaded into memory.
On the contrary, both TraVaG and the DDPM framework heavily rely on distinct ANN structures, with their training and application runtimes contingent upon various factors such as the model architecture, used library implementations, training schedules, and fine-tuned hyperparameters. Whereas for TraVaG, both the Autoencoder and the GAN need to be trained, the DDPM only comprises an untrained noise generator and a single denoising ANN. Due to batch processing, it's not imperative to load the entire event log into memory or GPU. On our hardware setup, once trained, TraVaG requires approximately 0.3 seconds to generate an artificial event log equivalent in size to BPIC-2012-App (comprising 13087 cases), while the DDPM generator takes about 7 seconds. The notable disparity in runtime stems from the iterative sampling approach employed by DDPMs, involving a sequence of denoising steps (300 in our implementation), wherein the trained network is applied at each step individually.
A more detailed investigation of training and application runtime can be found on Github\footnote{\scriptsize\url{https://github.com/wangelik/TraVaGen/blob/main/supplementary/Complexity_Supplementary.pdf}}.

As implied by the ANN-based generators of the TraVaG and DDPM frameworks described in Section \ref{sec:approach}, the internally learned variant distribution progressively enhances in accuracy with increasing size of the input event logs. Notably, performance improvements are observed with simpler distribution shapes and larger minimal case counts per variant.
This naturally prompts the inquiry into determining the minimum training data size requisite for model applicability. Our experiments reveal that this determination is contingent upon several factors, including the magnitude of DP, i.e., the induced noise during training, the architectural configuration of the model, and the trace uniqueness serving as a proxy for the distribution shape.
Given a specific event log, model complexity, and $(\epsilon, \delta)$ parameters, the limit can be ascertained through an iterative process of scaling down variant frequencies until model convergence becomes unattainable. If no convergence appears even for the original event log, likely explanations include either an ill-suited, often excessively intricate model architecture or a training data size below the threshold implied by current configurations.
In our experimentation with the Sepsis, BPIC-2013, and BPIC-2012-App datasets (see Section \ref{subsec:datasets}), the data utility analysis revealed that the threshold was reached with $\epsilon$ ranging between 0.01 and 0.001 for the investigated $\delta$ regime and an event log representing 1050 cases across 846 variants (Sepsis).

\section{Conclusion}\label{sec:conc}

With this work, we introduced two novel differentially private generative frameworks designed to facilitate the secure release of event data while ensuring quantified and guaranteed privacy.
Our methodologies have successfully demonstrated that both training a differentially private combination of autoencoders and GANs as well as employing anonymized DDPMs to synthesize anonymized event data from an underlying original variant distribution outperform current state-of-the-art variant anonymization techniques for strong privacy levels in the low ($\epsilon, \delta$) range or complex event data structures.
Our work on DP-based DDPM infrastructures is the first attempt to leverage privatized DDPMs on structured and high-dimensional tabular data.
Furthermore, both generative models offer unique advantages, including an outstanding resource-efficient execution, the absence of distorting noise thresholds, a general acceptance of continuous data streams, and zero fake variant generation.

Overall, these characteristics enable the underlying generative algorithms to work efficiently with complex event data and lower ranges for $\delta$, which is a unique feature to the best of our knowledge.
However, it is important to acknowledge that our frameworks entail a more intricate training process and privacy budget management compared to conventional methods, such as TraVaS \cite{rafiei_travas_short}.
Because of the DP-SGD mechanisms that rely on RDP, it is not possible to directly extract or insert the conventional DP parameters $(\epsilon, \delta)$ into the model training process, as explained in \cite{rdp}. Instead, we must employ the one-way procedure detailed in Section~\ref{subsec:privacy_acc}. Accordingly, this involves first obtaining the RDP parameters $(\epsilon, \alpha)$ based on the noise multiplier $\Phi$, sampling rate $q$, and the number of iterations $T$, and subsequently converting these $(\epsilon, \alpha)$ parameters into $(\epsilon, \delta)$.
As a result, guaranteeing specific privacy levels necessitates an iterative analysis of various ANN settings until a suitable configuration is identified. In future research, this dependency on privacy-related hyperparameters could be investigated more comprehensively and potentially integrated into a fully automated tuning strategy. Depending on the available computational resources, such an approach could then transform TraVaG and the DDPM engine into streamlined, parameter-free methods akin to TraVaS.

\bibliographystyle{splncs04}
\bibliography{references}

\end{document}